\documentclass[runningheads]{llncs}
\usepackage[T1]{fontenc}
\usepackage{amsfonts}
\usepackage{amsmath}
\usepackage{caption}
\usepackage{cleveref}
\usepackage{float}
\usepackage{subcaption}
\DeclareMathOperator*{\argmax}{arg\,max}

\usepackage{graphicx}
\begin{document}
%
\title{ConDL: Detector-Free Dense Image Matching}
%
%
\author{Monika Kwiatkowski\orcidID{0000-0001-9808-1133} \and
 Simon Matern\orcidID{0000-0003-3301-2203} \and
Olaf Hellwich\orcidID{0000-0002-2871-9266}}
\authorrunning{M. Kwiatkowski et al.}
%
\institute{Computer Vision \& Remote Sensing, \\ Technische Universität Berlin, \\ Marchstr. 23, Berlin, Germany
\email{\{m.kwiatkowski\},\{s.matern\},\{olaf.hellwich\}@tu-berlin.de}
}
\maketitle              
\begin{abstract}
In this work, we introduce a deep-learning framework designed for estimating dense image correspondences. Our fully convolutional model generates dense feature maps for images, where each pixel is associated with a descriptor that can be matched across multiple images. Unlike previous methods, our model is trained on synthetic data that includes significant distortions, such as perspective changes, illumination variations, shadows, and specular highlights. Utilizing contrastive learning, our feature maps achieve greater invariance to these distortions, enabling robust matching. Notably, our method eliminates the need for a keypoint detector, setting it apart from many existing image-matching techniques.

\end{abstract}
\section{Introduction}

Estimating correspondences is an crucial task in numerous computer vision problems. Accurate correspondences enable the estimation of various properties of the observed scene, such as camera motion and object geometry. Point matching across images is vital for tasks including structure from motion (SfM), image stitching, object tracking, image retrieval, and dense 3D reconstruction. \\ \\
\begin{figure}
    \centering
    \includegraphics[width=\textwidth]{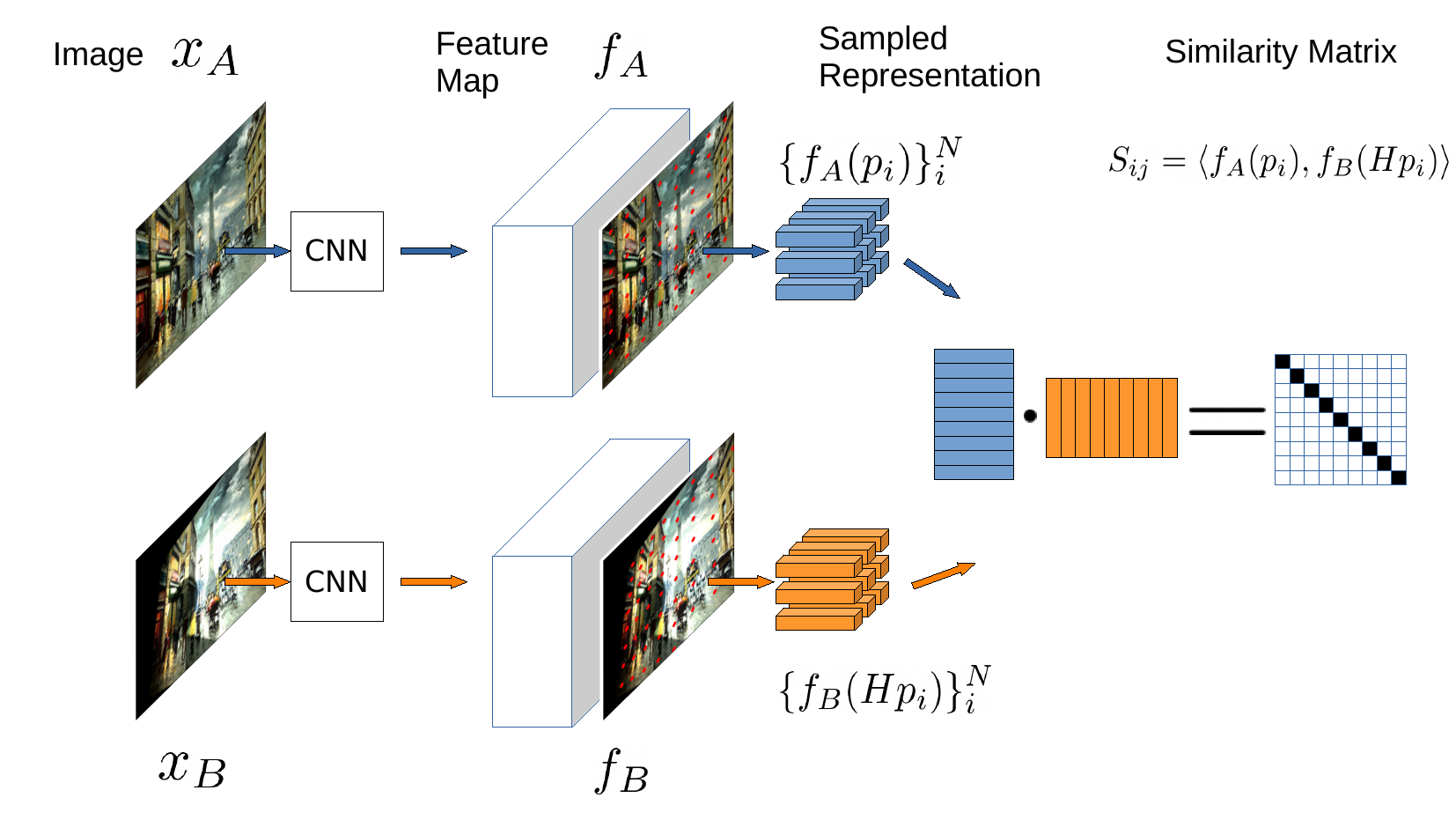}
    \caption{An illustration of the ConDL framework. Dense feature maps are extracted from two images. Keypoints are differentiably sampled from the feature maps. Matches are estimated from similarity scores by calculating pairwise dot-products.}
    \label{fig:condl}
\end{figure}
In this work, we present \textit{ConDL} (Contrastive Descriptor Learning), an advanced image-matching framework designed for computing dense correspondences. Leveraging synthetic data augmentations from SIDAR \cite{kwiatkowski2023sidar}, we generate training image-pairs under various perturbations, including perspective distortion, illumination changes, shadows, and occlusions. With ground truth homographies, we establish dense point correspondences, extracting dense image features using a CNN-based ResNet. Employing a contrastive learning approach, \textit{ConDL} learns a similarity metric to robustly match image features despite these perturbations. Unlike existing metric learning methods, \textit{ConDL} does not use a triplet loss or require a mining strategy for positive and negative samples. Instead, inspired by CLIP \cite{DBLP:journals/corr/abs-2103-00020}: Points are differentiably sampled from both feature maps \cite{DBLP:journals/corr/JaderbergSZK15}, we use a contrastive learning approach where points are differentiably sampled from both feature maps, and a similarity matrix of all correspondences is computed. The similarity score of matching features is maximized, while the score of incorrect matches is minimized. \\\\
To summarize, our method provides the following contributions:
\begin{itemize}
    \item \textbf{Dense Matching}: \textit{ConDL} establishes dense image matches across images. A fully convolutional ResNet estimates pixel-wise representations that can be matched.
    \item \textbf{Robustness:} By combining contrastive learning with image pairs featuring a wide variety of distortions, our model learns a more invariant representation.
    \item \textbf{Modularity}: Our framework consists of simple interchangeable components: (dense) feature extraction, differentiable sampling, and similarity matrix computation. It is adaptable to different models for feature extraction, and while we sample an equidistant grid, keypoints can be extracted using other strategies, such as classical keypoint detectors, while still utilizing \textit{ConDL's} robust descriptors.
\end{itemize}

\section{Related Work}

Many classical approaches rely on extracting sparse distinct keypoints with corresponding descriptors from a scene. In recent years, keypoint detectors and descriptors have been learned using deep learning approaches to increase the robustness of image matching.


\subsubsection{Metric Learning} Many approaches use metric learning to estimate a similarity between keypoints or patches directly \cite{zagoruyko2015learning,yi2016lift,choy2016universal,mishkin2018repeatability}. Siamese models extract features from a pair of images or patches, and a metric is estimated by minimizing the metric between positive samples and maximizing the metric between negative samples. \textit{ConDL} differs from these existing methods as it does not rely on patches. Our method is conceptually similar to Choy et al. (2016) \cite{choy2016universal}. The significant difference is that we use a different sampling strategy and different training loss for optimization. We do not use a triplet loss; instead, a similarity matrix is computed across all point pairs, and a cross-entropy loss is minimized for each keypoint, which requires optimization of all correspondences simultaneously. 

\subsubsection{Detector Learning:} In order to match images efficiently, many methods rely on sparse keypoints detection \cite{Barroso-Laguna2019ICCV,DBLP:journals/corr/abs-1712-07629,sarlin2020superglue}. Distinct features are extracted first before computing correspondences. Our method is detector-free. 

\subsubsection{Detector-Free Matching:} Recent advances in transformer architectures allow the computation of image matches using cross-attention\cite{sun2021loftr,chen2022aspanformer,wang2022matchformer}. These methods do not rely on detectors. Attention allows to learn the global context of all image features within each image and across images. In addition, local consistency of matches can be enforced using an optimal transport layer. Our method is similar to cross-attention insofar as we compute pairwise dot-products across images. However, we do not use any additional layers or processing to compute contextual features.

\subsubsection{Datasets:} Image matching methods often require ground truth correspondences for training. This limits the training often to SfM datasets \cite{MegaDepthLi18,Schops_2019_CVPR} and optical flow estimation \cite{Butler:ECCV:2012,Menze2015CVPR}. Since these methods usually depend on existing image-matching methods, the complexity of correspondences is limited by the data collection. Without any additional regularization, a learned feature extractor can only be as good as the image matching used during data collection. Our evaluations show that the training data has a significant influence on the performance and robustness of the method. Using the SIDAR pipeline\cite{kwiatkowski2023sidar}, we generate strong synthetic image distortions, which could not have been aligned with conventional image-matching methods. 

\section{Synthetic Data Augmentations}
\label{sec:sidar}
\begin{figure}
    \centering
     \begin{subfigure}[b]{0.24\textwidth}
         \centering
         \includegraphics[width=\textwidth]{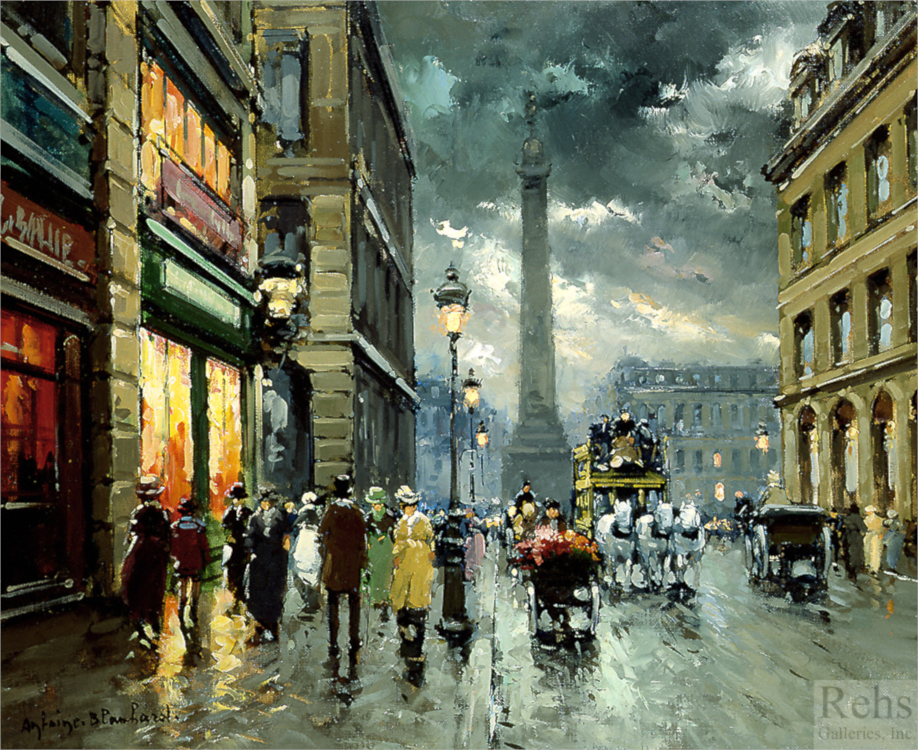}
         \caption{}
     \end{subfigure}
     \begin{subfigure}[b]{0.24\textwidth}
         \centering
         \includegraphics[width=\textwidth]{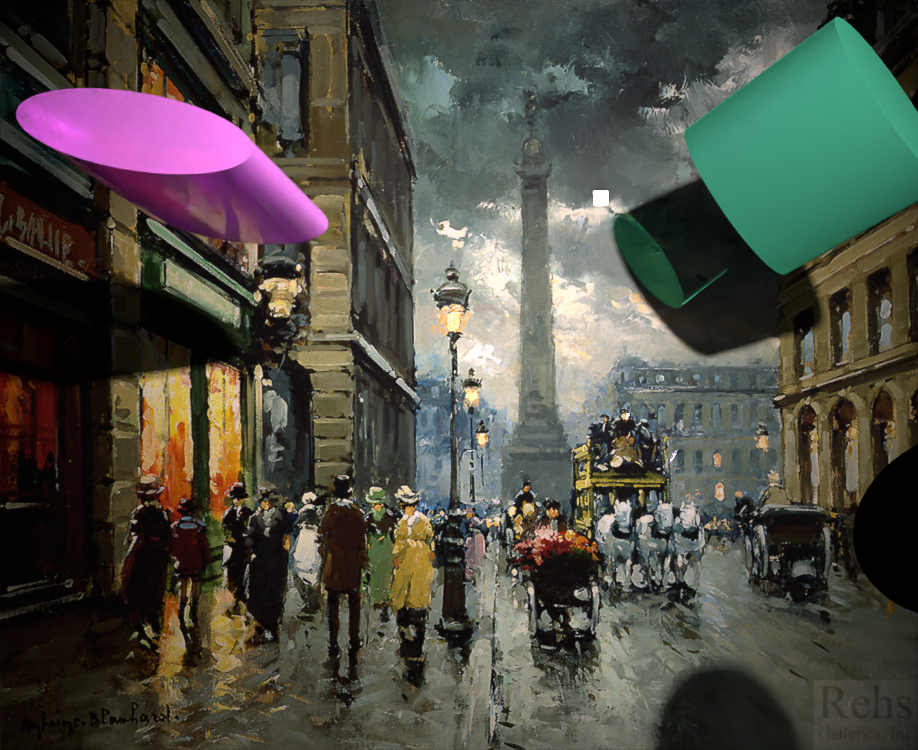}
         \caption{}
     \end{subfigure}
          \begin{subfigure}[b]{0.24\textwidth}
         \centering
         \includegraphics[width=\textwidth]{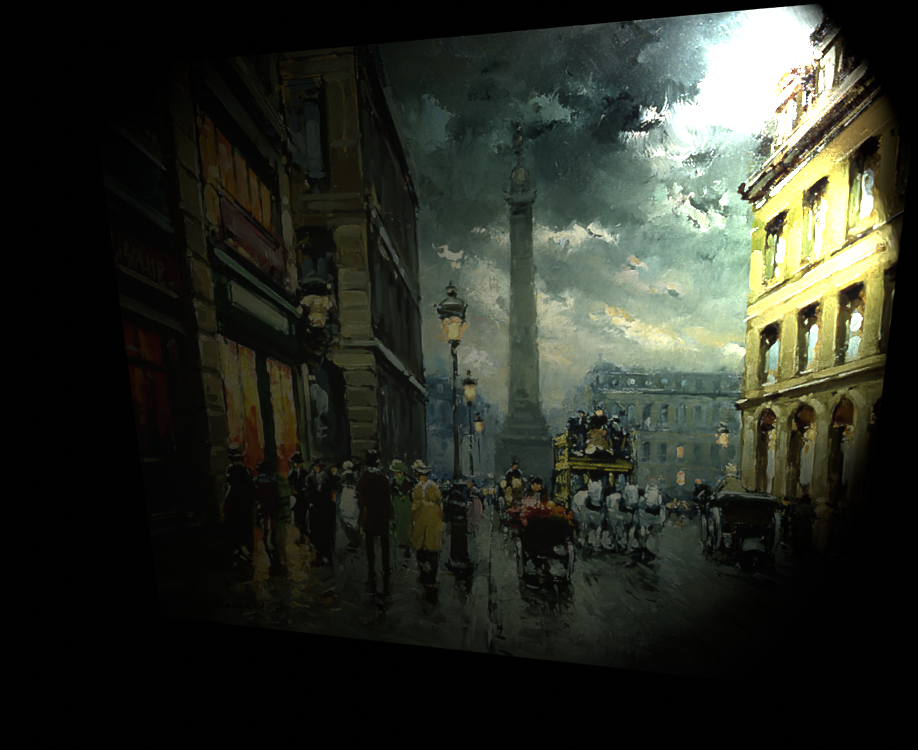}
         \caption{}
     \end{subfigure}
          \begin{subfigure}[b]{0.24\textwidth}
         \centering
         \includegraphics[width=\textwidth]{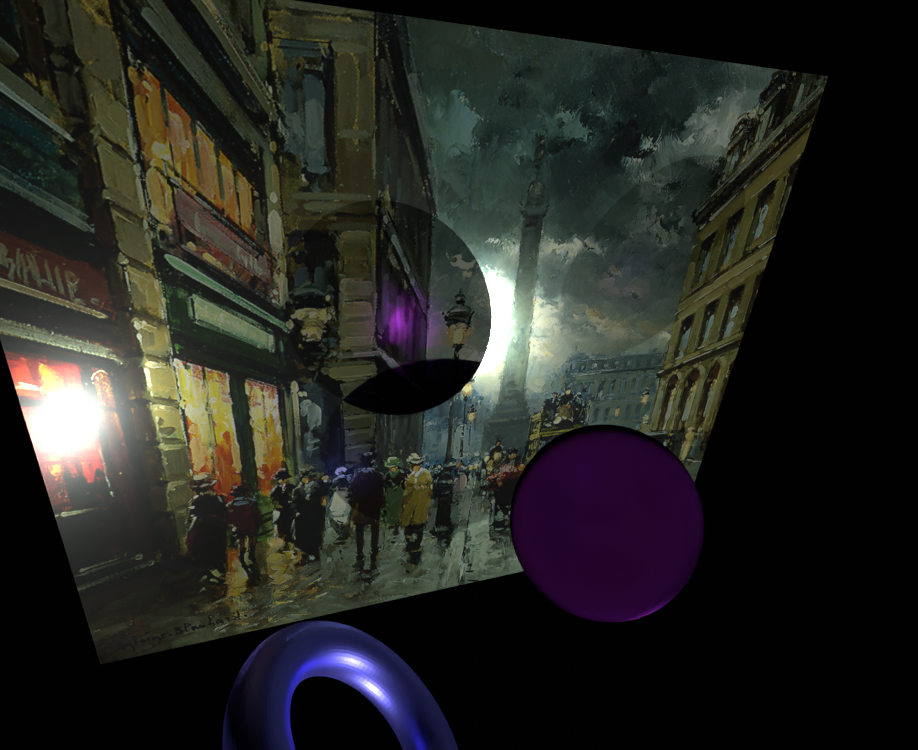}
         \caption{}
     \end{subfigure}
    \caption{(a) shows an input image and (b)-(d) show the created data augmentations.}
    \label{fig:Sidar}
\end{figure}

As illustrated in \cref{fig:Sidar}, we use SIDAR\cite{kwiatkowski2023sidar} to add image distortions to an arbitrary input image. The images contain strong illumination changes, occlusions, shadows, and perspective distortions. Since the relative position of cameras and 2D planes are known during data generation, image correspondences can be computed regardless of the complexity of the scene. We generate a dataset consisting of 50,000 image pairs for training and 4,000 image pairs for testing.

\section{Contrastive Dense Matching}

\Cref{fig:condl} illustrates the functionality of model \textit{ConDL}. Dense image features are computed from two images. Both feature maps are differentiable sampled using an equidistant grid and its perspective projection extracting descriptor for the corresponding keypoints. A pairwise dot-product is computed between all descriptors, resulting in a similar matrix of $S$. 
During training, we maximize the diagonal values, which describe ground truth correspondences, and minimize all remaining values. During inference, the row-wise and column-wise maxima of the similarity matrix are used to identify matches. \\
Although we fix the size of the sampling grid during training, the sampling rate can be changed arbitrarily for inference at the cost of increased memory consumption.
%
\subsection{Dense Feature Extraction}
Given two images $x_A,x_B\in \mathbb{R}^{3\times H\times W}$ we extract dense feature maps of the same resolution:
\begin{align}
    f_A = f_\theta (x_A) \in \mathbb{R}^{d\times H\times W}\\
    f_B = f_\theta (x_B) \in \mathbb{R}^{d\times H\times W}
\end{align}
We use a fully convolutional ResNet consisting of 10 residual blocks for the feature extraction $f_\theta$. 
Let $(p_i,p_j)$ with $p_i:=(\mathrm{x}_i,\mathrm{y}_i),p_j:=(\mathrm{x}_j,\mathrm{y}_j)$ be pair of corresponding pixels .
Each pixel in the original images has a corresponding descriptor:
\begin{align}
    f_A(p_i),f_B(p_j) \in \mathbb{R}^{d}
\end{align}

In order to find pixel correspondences $(p_i,p_j)$ during inference we maximize the dot-product:
\begin{align}
   p_j := \argmax_{p_k} \langle f_A(p_i),f_B(p_k)\rangle
\end{align}

This concept is also similar to the cross-attention of transformers \cite{vaswani2017attention}, which also computes the dot-product between two sequences of tokens. 

\subsection{Differentiable Sampling}
\begin{figure}
    \centering
    \begin{subfigure}[b]{0.45\textwidth}
         \centering
         \includegraphics[width=\textwidth]{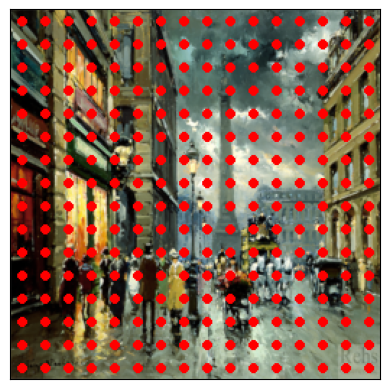}
         \caption{Equidistantly sampled grid}
     \end{subfigure}
    \begin{subfigure}[b]{0.45\textwidth}
         \centering
         \includegraphics[width=\textwidth]{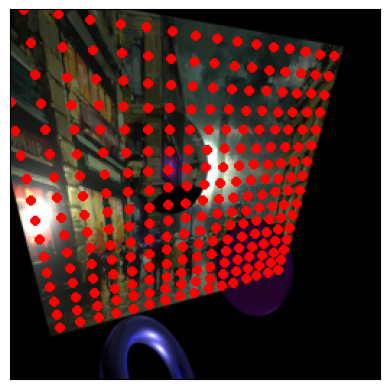}
        \caption{Projected grid points}
     \end{subfigure}
         
    \caption{Illustration of sampled point correspondences.}
    \label{fig:grid-sample}
\end{figure}

\begin{figure}
    \centering
\begin{subfigure}[b]{0.45\textwidth}
         \centering  \includegraphics[width=\textwidth]{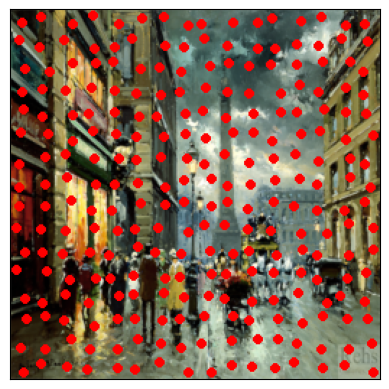}
         \caption{Equidistant grid with noise}
     \end{subfigure}
    \begin{subfigure}[b]{0.45\textwidth}
         \centering
\includegraphics[width=\textwidth]{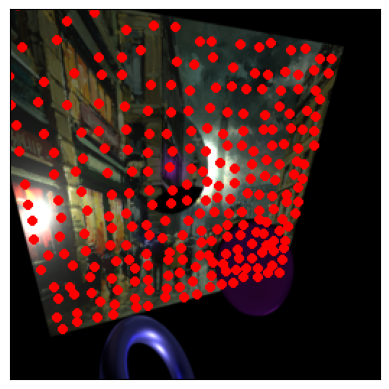}
        \caption{Projected grid points with noise}
     \end{subfigure}
    \caption{Illustration of sampled point correspondences with added noise.}
    \label{fig:grid-sample2}
\end{figure}

As described in \Cref{sec:sidar}, our training data consists of image pairs with perspective distortions. In order to learn robust representations, the features need to be aligned first; the dataset provides ground truth homography and allows the extraction of pixel-wise correspondences. However, matching all pixels against each other has an $\mathcal{O}(n^2)$ memory complexity. Instead, we extract a much sparser grid of points. \\\\
We create a uniform sample grid of points 
$\{p_i\}_i^N \subset [0, W-1]\times [0,H-1]$, where $W$ and $H$ are the image width and image height respectively. Given the known homography $\mathcal{H}$, we project the grid points into the other image resulting in a perspective projection of the grid $\{\mathcal{H}p_i\}_i^N$. In order to avoid overfitting due to repeatedly sampling the exact same points, we add noise to our initial grid points. 
\Cref{fig:grid-sample,fig:grid-sample2} illustrates our sampling method. 

We utilize the differentiable image sampling method introduced by Jaderberg et al. (2015) \cite{DBLP:journals/corr/JaderbergSZK15}. Let $U \in \mathbb{R}^{C\times H \times W}$ be a feature map and  $G \in \mathbb{R}^{2\times H' \times W'}$ a sampling grid. Each grid point $p_i$ contains the normalized pixel location $(\mathrm{x}_i,\mathrm{y}_i)$ in the feature map $U$:\[(\mathrm{x}_i,\mathrm{y}_i) = G(p_i) \in [-1,+1]^2 \] 
A new feature map $V \in \mathbb{R}^{C\times H' \times W'}$ can be differentiable computed by copying the values from $U$ at position $(\mathrm{x}_i,\mathrm{y}_i)$ to the grid location $p_i$. Using bilinear interpolation the sampled feature value $V(p_i)$ is computed as:
\begin{align}
    V(p_i)^c = \sum_n^H \sum_m^W U_{n,m}^c \max\left(0, 1 - \left| \mathrm{x}_i - \frac{m}{W} +0.5 \right|\right) \max\left(0, 1-\left|\mathrm{y}_i - \frac{n}{H} + 0.5 \right|\right)  \label{eq:grid-sample}
\end{align}
\\
Given two feature maps $f_A,f_B \in \mathbb{R}^{C\times H \times W}$, the grid $\{p_i\}_{i=1}^N$ and its projection $\{\mathcal{H}p_i\}_{i=1}^N$ we extract the keypoints' descriptors $f_A(p_i),f_B(\mathcal{H} p_i) $  as described in \cref{eq:grid-sample}. 

\subsection{Contrastive Loss}

Given a set of sampled descriptors $\{f_A(p_i)\}_{i=1}^N$
from image $x_A$ and the matching descriptors $\{f_B(\mathcal{H}p_i)\}_{i=1}^N$ from image $x_B$ we compute a similarity matrix $S \in \mathbb{R}^{N\times N}$ using pairwise dot-products:
\begin{align}
    S_{ij} = \langle f_A(p_i),f_B(\mathcal{H}p_j)\rangle
\end{align}
Note that only the diagonal entries $S_{ij}$ describe scores of correct matches. We follow the approach of CLIP\cite{DBLP:journals/corr/abs-2103-00020} and compute a row-wise and column-wise softmax:
\begin{align}
   \text{Row-wise Softmax:}&& p_A(i,j) &= \dfrac{\exp{\left(S_{ij}\right)}}{\sum_{k=1}^N \exp{\left(S_{ik}\right)}} \\
  \text{Column-wise Softmax:}&& p_B(i,j) &= \dfrac{\exp{\left(S_{ij}\right)}}{\sum_{k=1}^N \exp{\left(S_{kj}\right)}} 
\end{align}

The row $p_A(i,:)$ describes the matching distribution over all keypoints in image $x_B$. The column $p_B(:,j)$ describes the matching distribution over all keypoints in image $x_A$, respectively. We can define the matching as a classification problem for each keypoint:
\begin{align}
    i \overset{!}{=} \argmax_k p_A(i,k) ~~ \forall i=1,\dots,N\\
    i \overset{!}{=} \argmax_k p_B(k,i) ~~\forall i=1,\dots,N
\end{align}

A cross-entropy is computed for each row and each column of $p_A$ and $p_B$, respectively.
\begin{align}
    L_A = \dfrac{1}{N}\sum_{i=1}^N \log(p_A(i,i)) \\
    L_B = \dfrac{1}{N}\sum_{i=1}^N \log(p_B(i,i)) 
\end{align}
The final loss used for training is the total average overall matches:
\begin{align}
    L = \dfrac{L_A + L_B}{2}
\end{align}

Unlike other learned image matching methods\cite{choy2016universal,sun2021loftr,sarlin20superglue}, our training does not require nearest neighbor searches, analyzing patches, or complex mining for positive and negative samples. All descriptors are optimized against each other.
However, the computation of the similarity matrix creates a bottleneck in our framework due to memory consumption.
Since our framework is flexible in terms of the number of sampled points, in future work, we would like to evaluate the effect of the sampling rate on training and generalization.  

\subsection{Training}
For feature extraction, we use a ResNet with ten residual blocks, batch normalization, and 128 feature channels. We train on an NVIDIA RTX A6000 with 48 GB memory. A batch size of $16$ is used with a sampling grid of size $16 \times 16$. We use an Adam optimizer with a learning rate of $1e-3$ and default parameters $(\beta_1,\beta_2) = (0.9, 0.999)$ and $\epsilon=1e-8$. Training for 500 epochs on the given setup takes $\sim 60$ hours.

\section{Evaluation}

Using SIDAR \cite{kwiatkowski2023sidar}, we generate a test set of 4000 image pairs with corresponding ground truth homographies. An image pair consists of one undistorted image, and its distorted version contains strong illumination changes, perspective distortions, occlusions, and shadows. We evaluate various classical and state-of-the-art image-matching methods on the test set. \\\\
Our goal is to estimate the reliability and quality of each matching method. For each image pair, we compute point correspondences. From the estimated point pairs, we compute the homography using RANSAC.  
We evaluate the estimation of the homography by computing the mean corner error (MCE):
$$MCE(H,H') = \sum_{i=1}^4 \lVert Hx_i-H'x_i   \rVert_2$$
Where $x_i$ describes the corners of the image, this gives an estimation of the quality of the matches. The more accurate correspondences, the closer we get to the ground truth homography. 
Furthermore, we evaluate the individual matches $p_i\leftrightarrow p_i'$ by computing the reprojection error:
$$ L(p,p') = \lVert Hp_i-p_i'   \rVert_2  $$
The error is measured in pixels.
We count the number of inliers based on various thresholds $t \in \{ 0.1, 1, 10\}$. We do not consider correspondences with a larger error since the likelihood increases that they are outliers, and their reprojection errors are due to chance and not matching accuracy.  \\\\
We evaluate our method using various sampling rates. In the following, we use \textbf{ConDL 2px}, \textbf{ConDL 4px}, etc., to describe a sampling rate of every 2 pixels, and 4 pixels, respectively.
OpenCV \cite{opencv_library}, and Kornia \cite{riba2020kornia} provide many classical and state-of-the-art keypoint detectors and image descriptors.
For the classical/unsupervised methods we use SIFT \cite{lowe1999object}, ORB \cite{rublee2011orb}, AKAZE, and BRISK \cite{tareen2018comparative}. 
For supervised methods, we use LoFTR \cite{sun2021loftr} and Superglue \cite{sarlin2020superglue}. Kornia also provides various combinations of keypoint detectors (GFFT \cite{shi1994good} and KeyNet \cite{Barroso-Laguna2019ICCV}) and descriptors (AffNet and HardNet \cite{mishkin2018repeatability}). LoFTR has weights for indoor scenes (LoFTR-i) and outdoor scenes (LoFTR-o). 
\subsection{Quantitative Results}
\begin{figure}
    \centering
\includegraphics[width=0.9\textwidth]{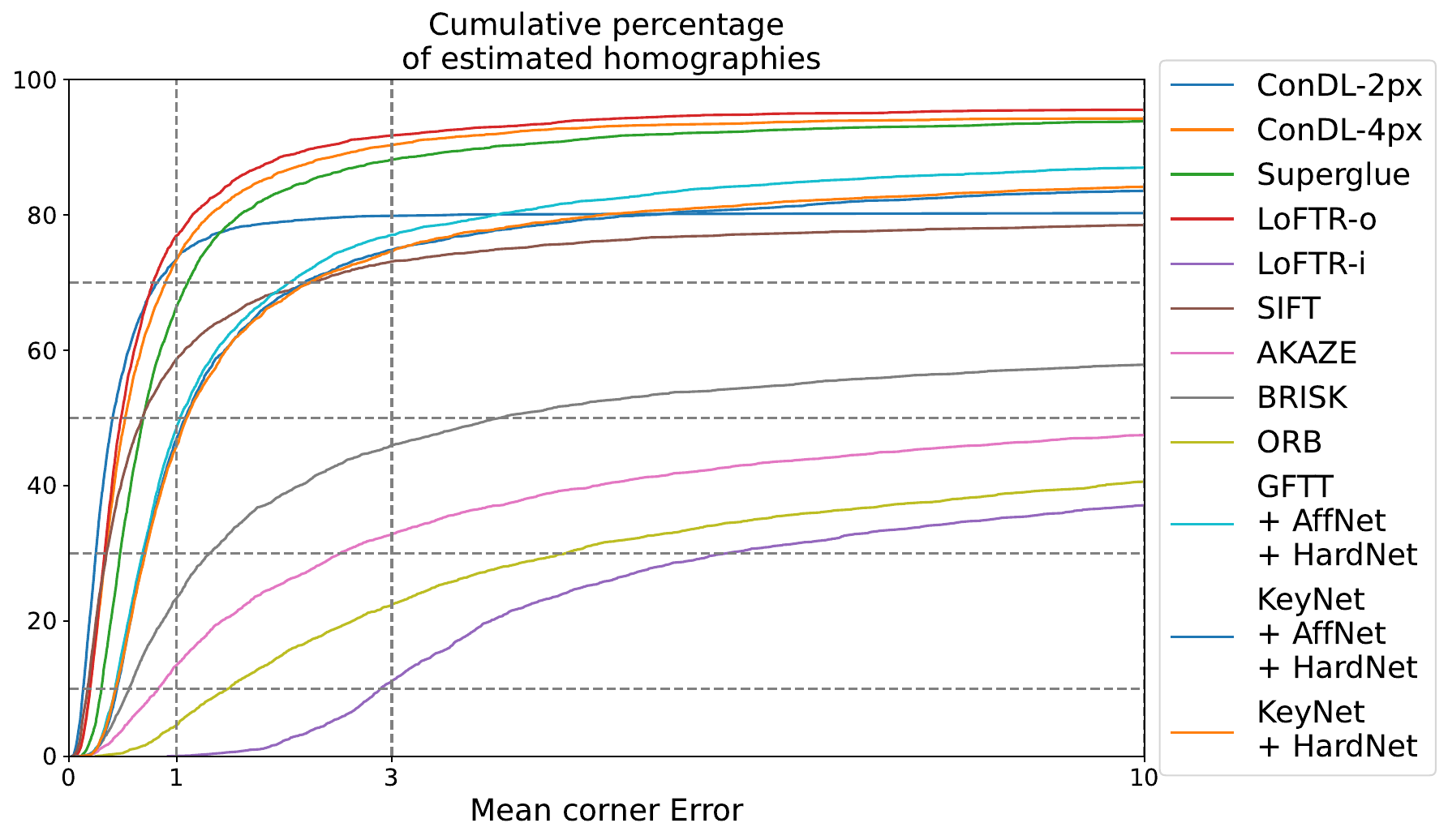}
    \caption{The graphs show the cumulative percentage of
estimated homographies below a given Mean Corner Error.}
    \label{fig:mce-cum}
\end{figure}

\begin{figure}
    \centering
\includegraphics[width=0.9\textwidth]{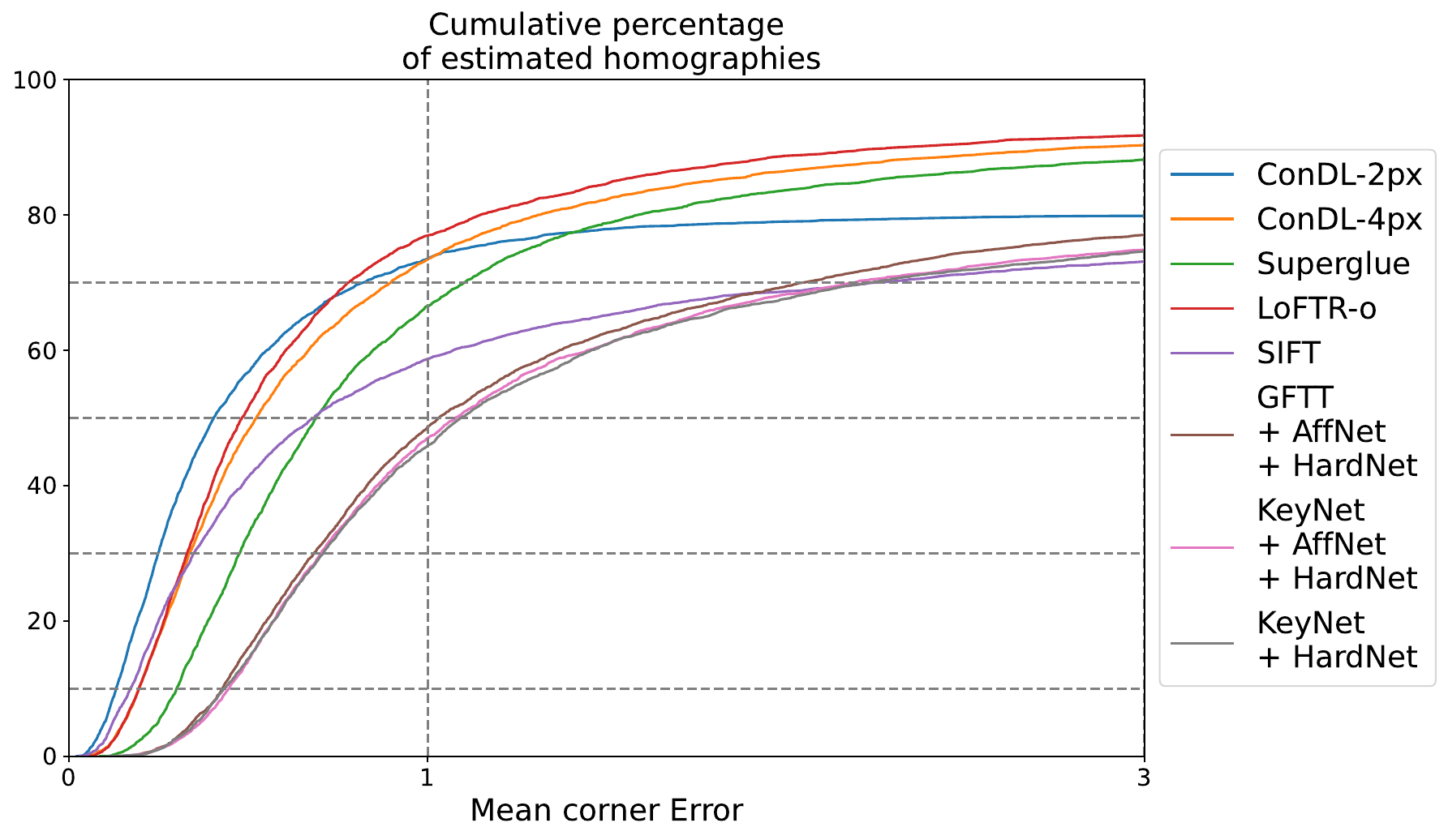}
    \caption{The graph shows the cumulative distribution for homography estimations close to subpixel accuracy.}
    \label{fig:mce-zoom}
\end{figure}

\begin{figure}
    \centering
    \begin{subfigure}[b]{0.45\textwidth}
         \centering
         \includegraphics[width=\textwidth]{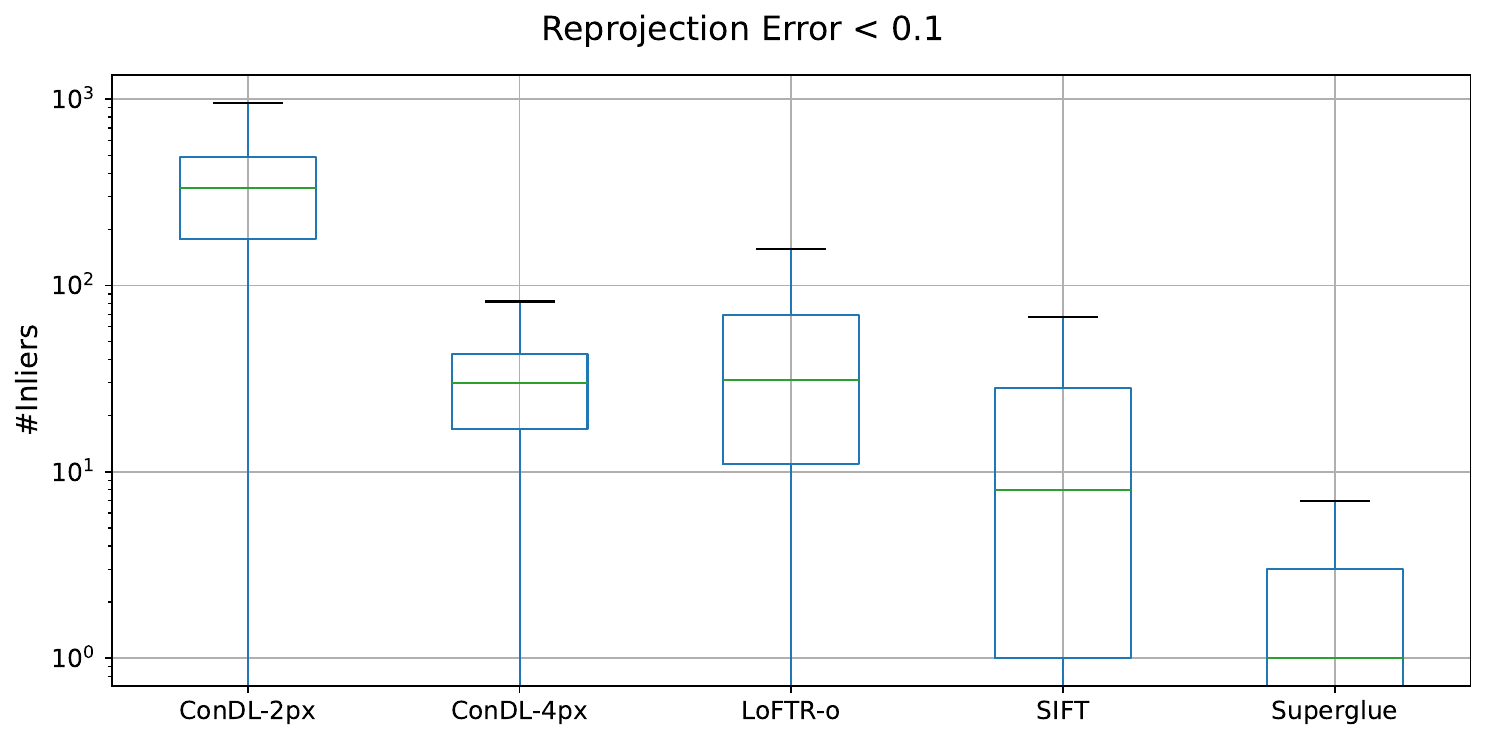}
         \caption{}
     \end{subfigure}
     \begin{subfigure}[b]{0.45\textwidth}
         \centering
\includegraphics[width=\textwidth]{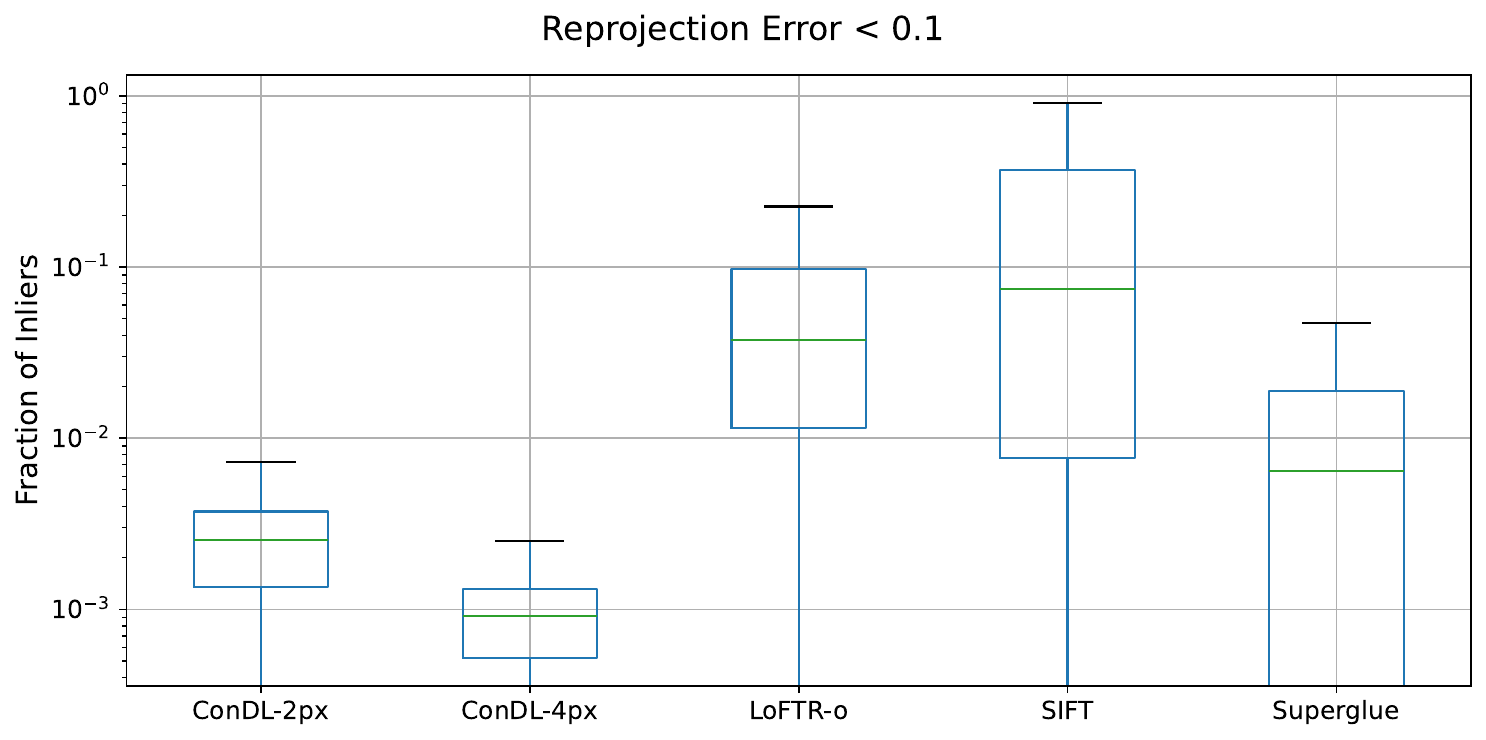}
         \caption{}
     \end{subfigure}
    \caption{(a) Number of correct matches with a reprojection error < 0.1. (b) The corresponding fraction of inliers relative to all matches.}
    \label{fig:0-1}
\end{figure}

\begin{figure}
    \centering
    \begin{subfigure}[b]{0.45\textwidth}
         \centering
         \includegraphics[width=\textwidth]{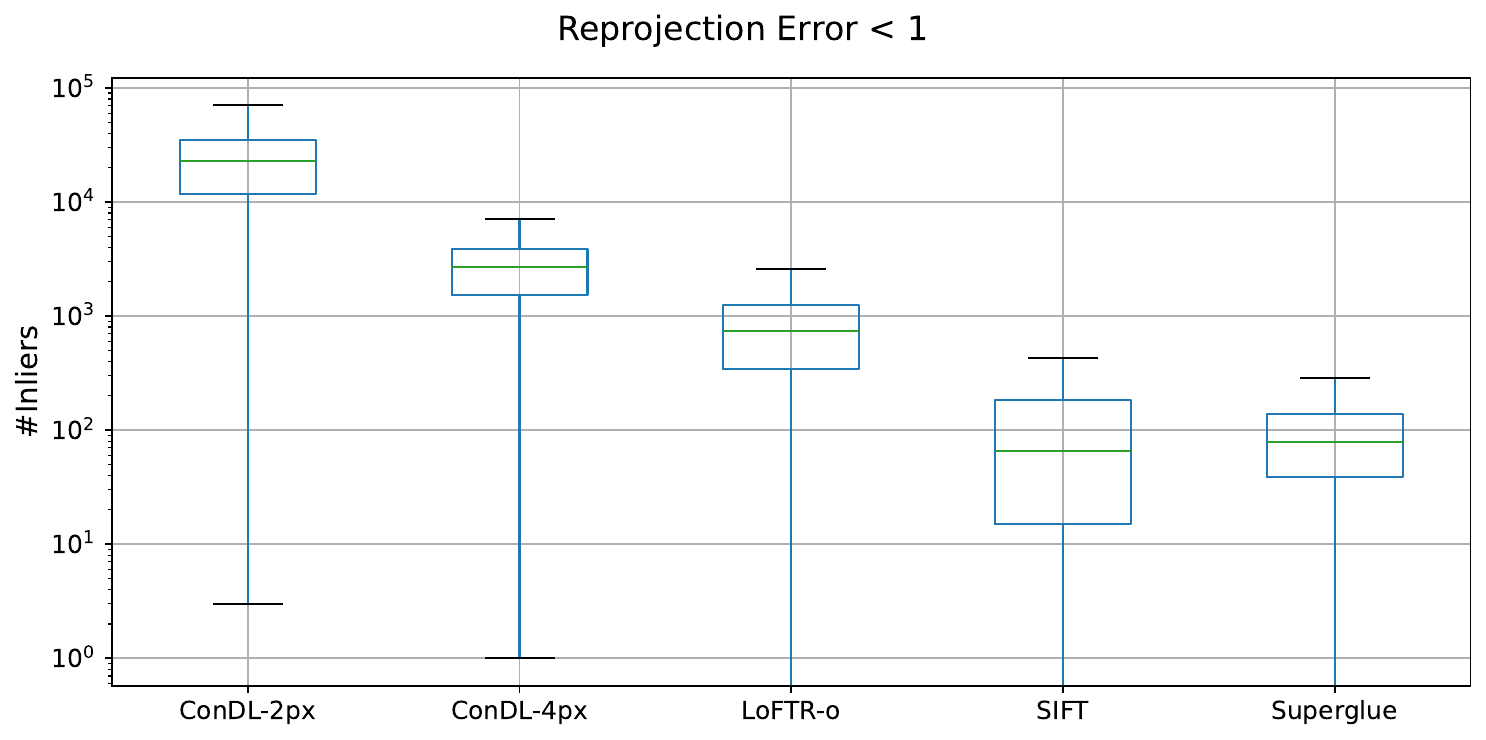}
         \caption{}
     \end{subfigure}
     \begin{subfigure}[b]{0.45\textwidth}
         \centering
\includegraphics[width=\textwidth]{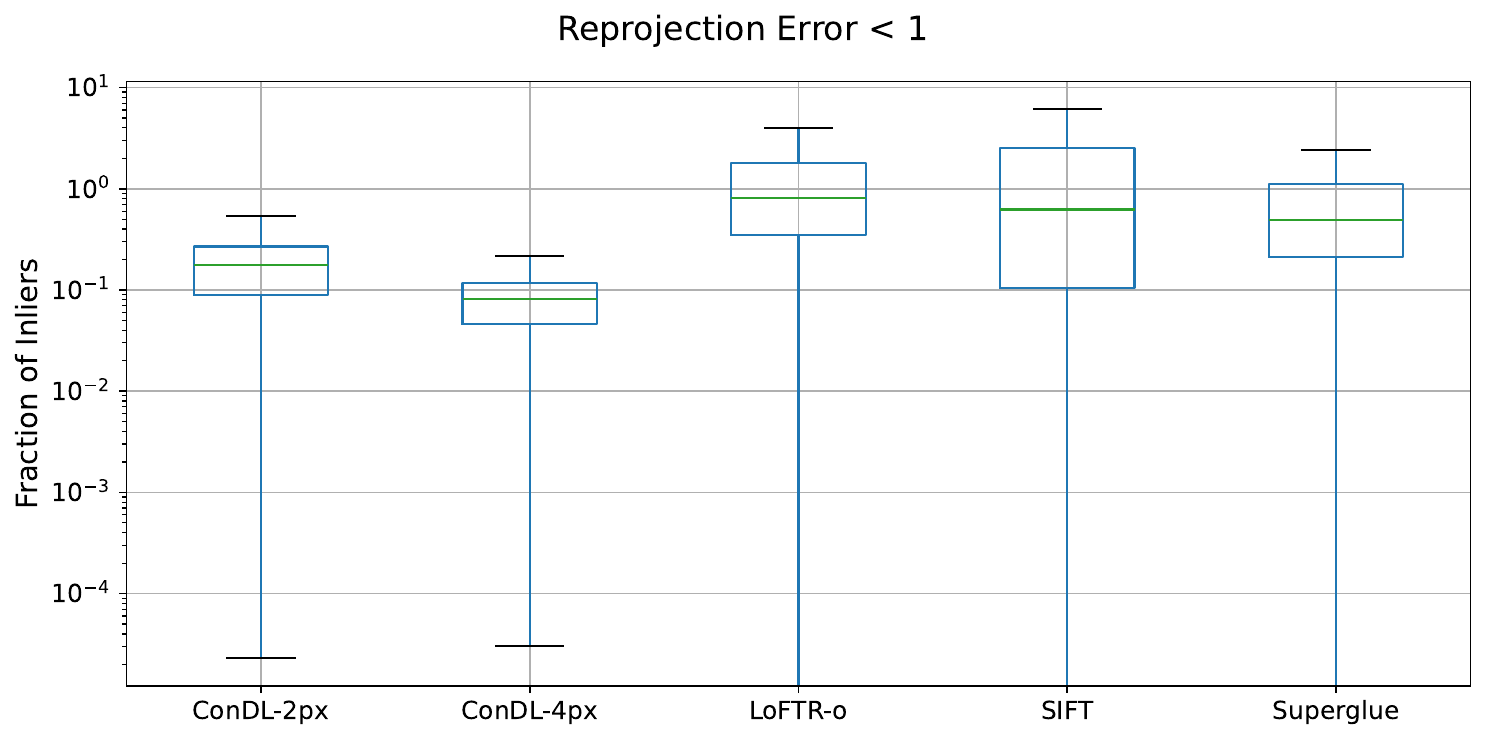}
         \caption{}
     \end{subfigure}
    \caption{(a) Number of correct matches with a reprojection error < 1. (b) The corresponding fraction of inliers relative to all matches.}
    \label{fig:1}
\end{figure}

\Cref{fig:mce-cum,fig:mce-zoom} illustrate the quality and robustness of the homography estimation using various image matchers. 
\begin{figure}
    \centering
\includegraphics[width=0.9\textwidth]{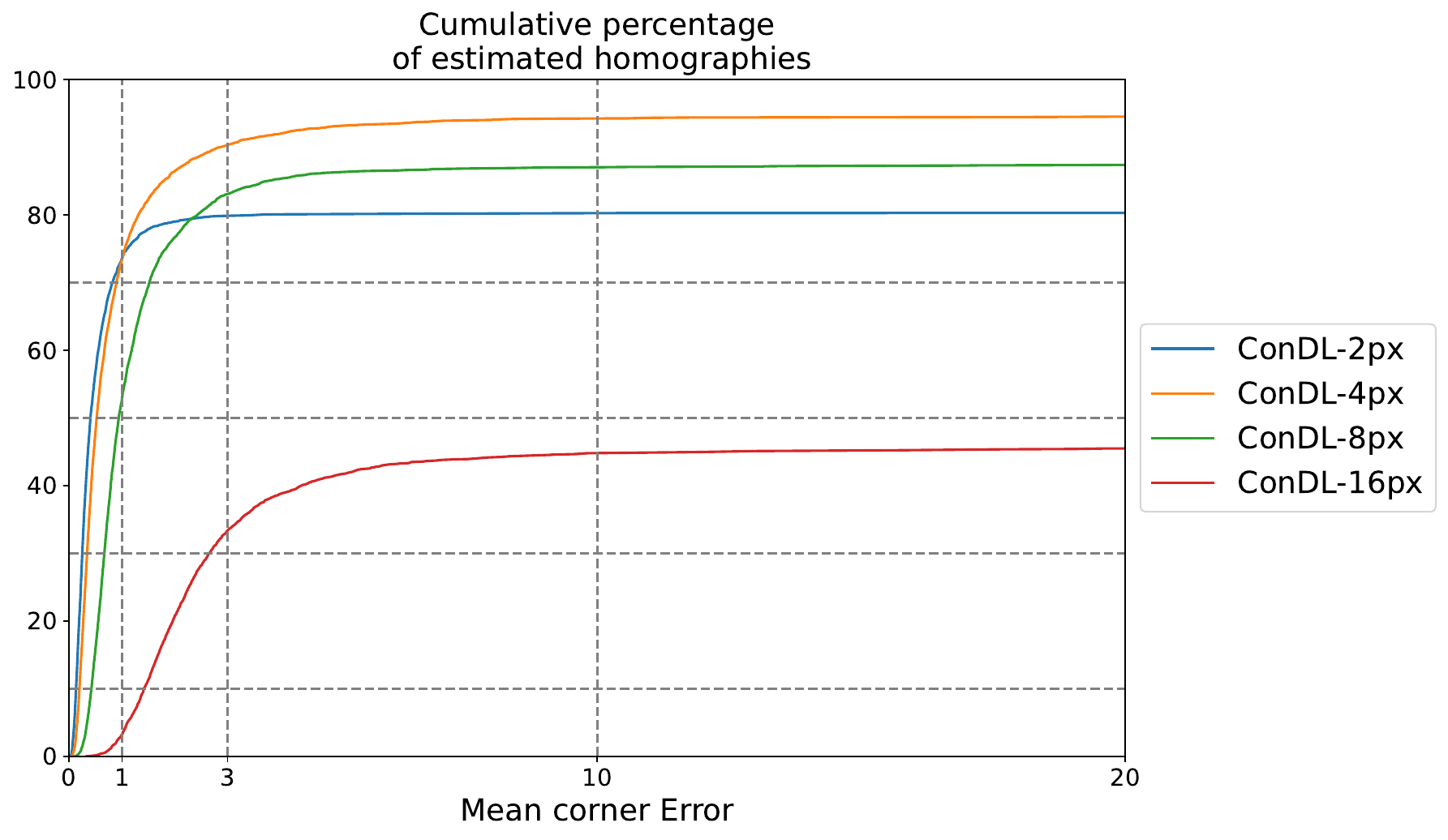}
    \caption{A comparison of ConDL with varying sampling rates}
    \label{fig:mce-condl}
\end{figure}
The results confirm the original SIDAR experiments\cite{kwiatkowski2023sidar}: trained descriptors outperform conventional methods. SIFT performs comparably well to the trained methods. The results show that \textit{ConDL}, with the highest sampling rate, has the most estimations with subpixel accuracy, although the performance is stagnating. This is due to the high sampling rate, which leads to many false-positive matches. The low ratio of inliers to outliers requires more iterations during RANSAC. This is also confirmed in \cref{fig:0-1,fig:1}. 
\textit{ConDL-4px}, on the other hand, is more robust, works comparatively well with LoFTR, and outperforms Superglue. \Cref{fig:mce-condl} shows the effect of different sampling rates. A high sampling rate increases the quality of the correspondences at the cost of robustness. Increasing the number of RANSAC iterations would improve robustness but increase computational cost, whereas increasing the sampling rate can also lead to significant degradation in performance.
In our current implementation, we do not discard any correspondences. Each keypoint is matched according to the largest similarity score. Using additional thresholding, it would be possible to discard ambiguous matches. \\
The diverging results of LoFTR-indoor and LoFTR-outdoor also showcase the effect of the training set and the learned biases.

\begin{figure}
    \centering
    \begin{subfigure}[b]{0.45\textwidth}
         \centering
         \includegraphics[width=\textwidth]{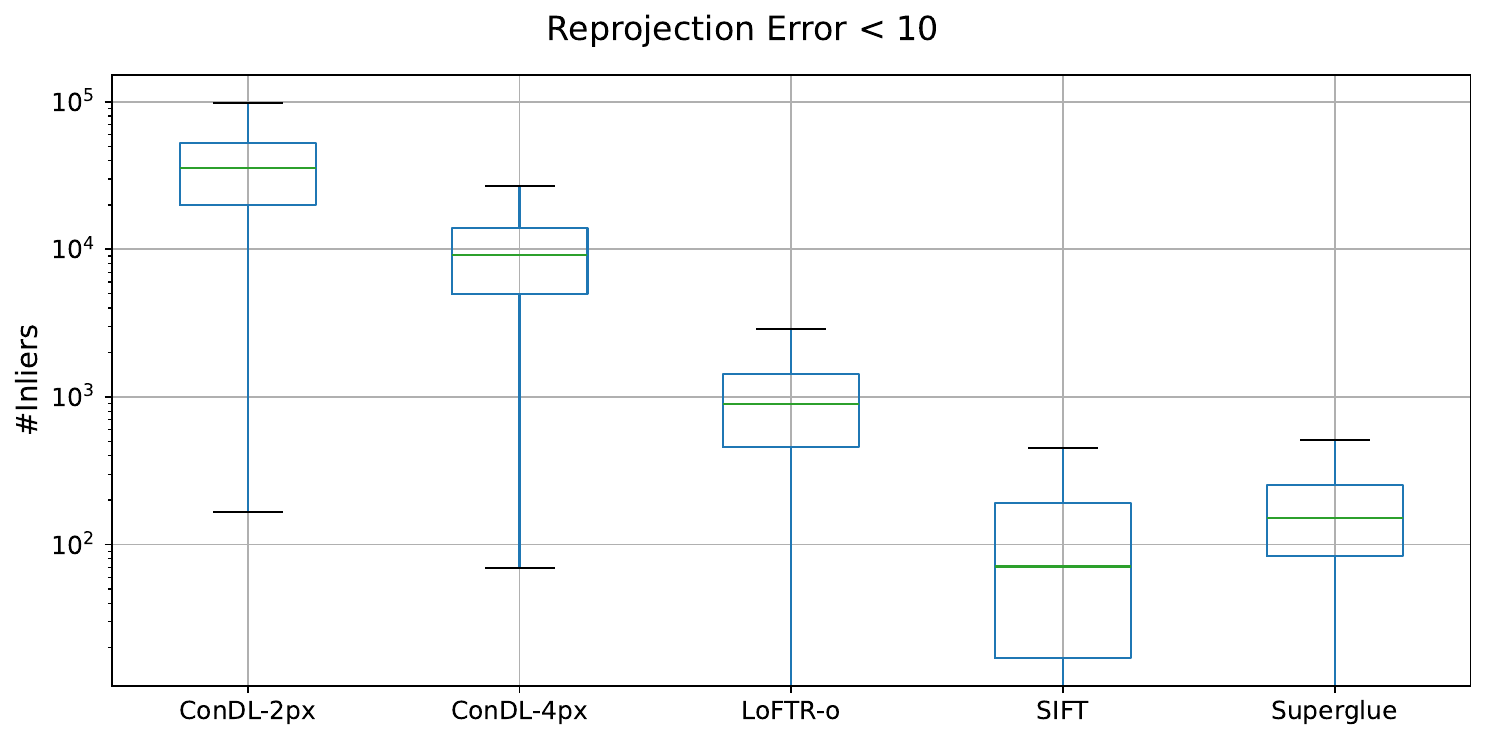}
         \caption{}
     \end{subfigure}
     \begin{subfigure}[b]{0.45\textwidth}
         \centering
\includegraphics[width=\textwidth]{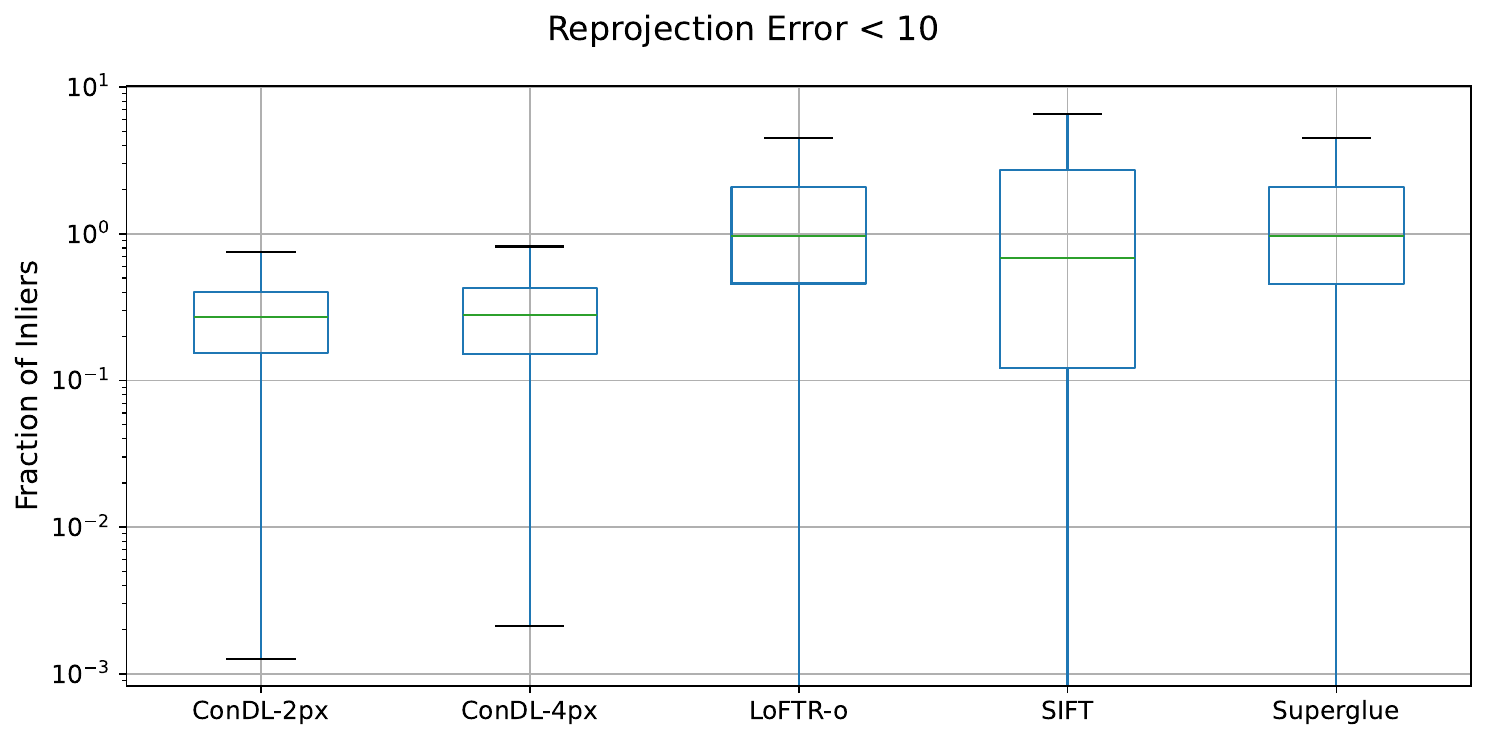}
         \caption{}
     \end{subfigure}
    \caption{(a) Number of correct matches with a reprojection error < 10. (b) The corresponding fraction of inliers relative to all matches.}
    \label{fig:10}
\end{figure}
\subsection{Qualitative Results}

\begin{figure}
    \centering
    \includegraphics[width=0.7\linewidth]{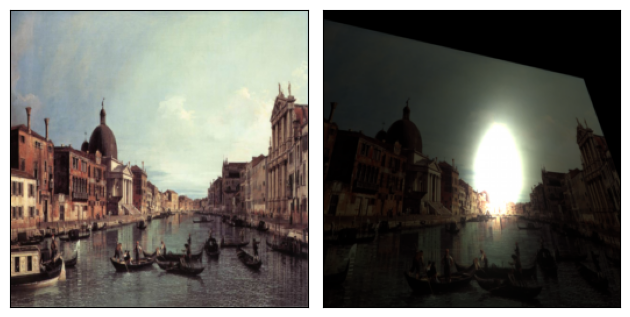}
    \includegraphics[width=0.45\linewidth]{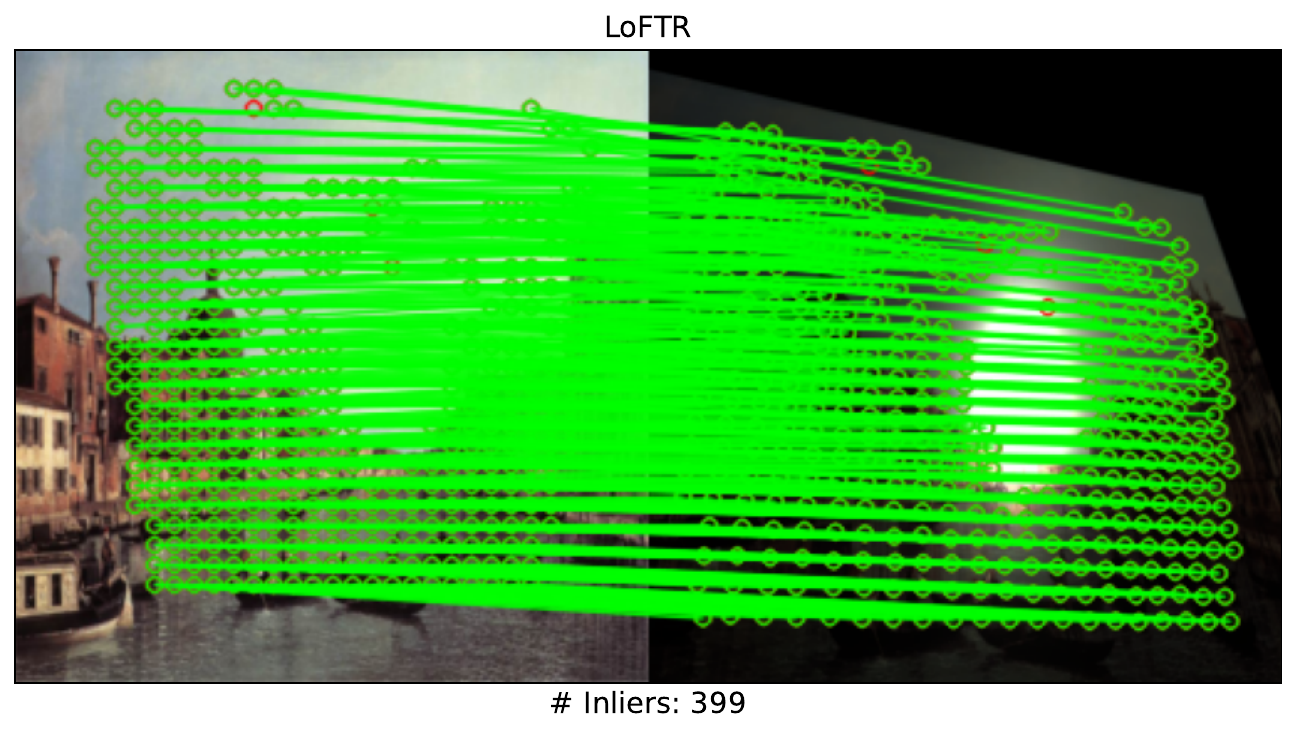}
   \includegraphics[width=0.45\linewidth]{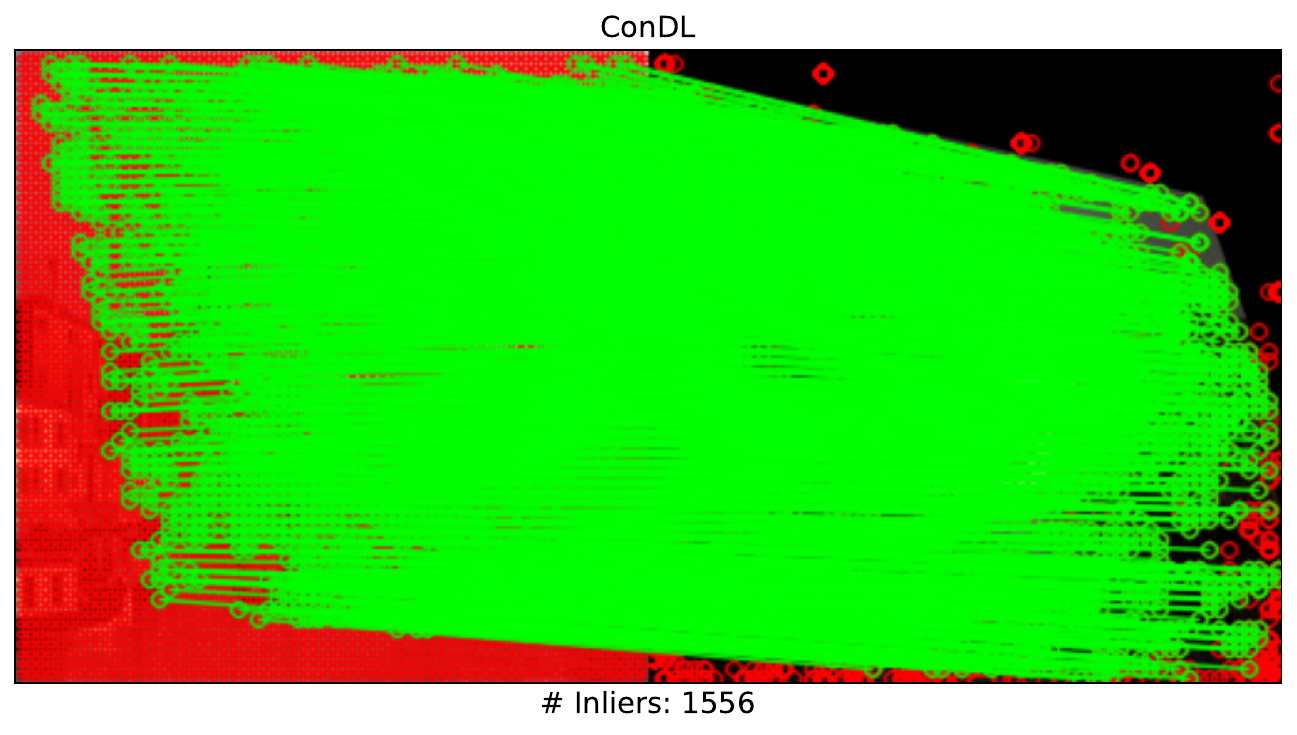}
    \caption{The first row shows an image pair under strong illumination changes. The second row illustrates the identified matches of LoFTR and ConDL. Green lines describe matches with a low reprojection error. Red circles describe keypoints with incorrect matches.}
    \label{fig:loftr-vs-condl}
\end{figure}

\Cref{fig:loftr-vs-condl} illustrates the matches found by \textit{ConDL} and LoFTR. \textit{ConDL} finds much more numerous and dense inliers, but there are many incorrect matches. LoFTR, on the other hand, only has a few incorrect matches. Almost all matches are inliers. The results show that learnable descriptors can be robustly trained to find matches even under very strong perturbations.
Although our dataset has a large variety of different scenes and distortions, our model does not yet generalize well to other datasets. 

\section{Conclusion}

In this work, we developed \textit{ConDL}, a robust image matching framework. \textit{ConDL} stands for \textit{Contrastive Descriptor Learning}. Our approach uses synthetic data augmentations for training, enabling the learning of image descriptors under arbitrarily complex perturbations. This is a significant advancement over many existing methods that rely on datasets derived from Structure-from-Motion (SfM) techniques, which often lack diverse noise and varied scenes. \\ 
\textit{ConDL} allows the computation of dense feature maps without relying on a keypoint detector. By using a differentiable grid sampler, we can explicitly control the sparsity of key points. Unlike state-of-the-art methods, such as LoFTR and Superglue, \textit{ConDL} does not rely on the relative positions of key points, resulting in more robust matching. Our evaluations demonstrate that \textit{ConDL} achieves performance comparable to state-of-the-art methods on our synthetic dataset.
In future work, we aim to train and evaluate \textit{ConDL} on additional datasets to further enhance its generalization capabilities.

%
%
%
 \bibliographystyle{splncs04}
\bibliography{references}

\end{document}